\documentclass[journal]{IEEEtran}

\usepackage{cite}
\usepackage{amsmath,amssymb,amsfonts}
\usepackage{graphicx}
\usepackage{textcomp}
\usepackage[utf8]{inputenc} 
\usepackage[T1]{fontenc}

\usepackage{bm}

\usepackage{accents}
\usepackage{algorithm}
\usepackage{algpseudocode}

\usepackage{caption}

\usepackage{eqparbox}
\usepackage{physics}
\usepackage{lipsum}
\usepackage{xspace} 

\usepackage{tabularx}
\usepackage{array, tabularx, boldline, dcolumn, booktabs}
\usepackage{float}
\usepackage{tablefootnote}
\usepackage{soul}
\usepackage[normalem]{ulem}
\usepackage{multirow}

\usepackage{mathtools}

\def \telcomodel           {\textit{TelcoInsight}}

\newcolumntype{L}[1]{>{\raggedright\let\newline\\\arraybackslash\hspace{0pt}}m{#1}}
\newcolumntype{C}[1]{>{\centering\let\newline\\\arraybackslash\hspace{0pt}}m{#1}}
\newcolumntype{R}[1]{>{\raggedleft\let\newline\\\arraybackslash\hspace{0pt}}m{#1}}

\begin{document}
\title{LLM-Augmented Knowledge Base Construction For Root Cause Analysis}

\author{Nguyen~Phuc~Tran,
        Brigitte~Jaumard,~\IEEEmembership{Senior,~IEEE,}
        Oscar~Delgado,~\IEEEmembership{Member,~IEEE}

        Tristan~Glatard,
        Karthikeyan~Premkumar,
        and Kun~Ni
\thanks{This work was supported by NSERC (under project ALLRP 566589-21) and InnovÉÉ (INNOV-R program) through the partnership with Ericsson and ECCC. We are grateful to Qinan Qi, Adel Larabi and Jennie Diem Vo at Ericsson-GAIA, Montréal for their support throughout this project.}

\thanks{N. P. Tran, B. Jaumard and T. Glatard are with the Department of Computer Science and Software Engineering, Concordia University, Montréal, Québec, Canada.}%
\thanks{O. Delgado is with École de Technologie Supérieure, Montréal, Québec, Canada.}%
\thanks{K. Premkumar, and K. Ni are with GAIA-Ericsson Montréal, Canada.}%
}

\markboth
{Phuc et al.: LLM-Augmented Knowledge Base Construction For Root Cause Analysis}
{Phuc et al.: LLM-Augmented Knowledge Base Construction For Root Cause Analysis}

\maketitle

\begin{abstract}
Communications networks now form the backbone of our digital world, with fast and reliable connectivity.
However, even with appropriate redundancy and failover mechanisms, it is difficult to guarantee “five 9s” (99.999\%) reliability, requiring rapid and accurate root cause analysis (RCA) during outages.
In the event of an outage, rapid and accurate RCA becomes essential to restore service and prevent future disruptions.

This study evaluates three Large Language Model (LLM) methodologies — Fine-Tuning, RAG, and a Hybrid approach — for constructing a Root Cause Analysis (RCA) Knowledge Base from support tickets.
We compare their performance using a comprehensive suite of lexical and semantic similarity metrics.
Our experiments on a real industrial dataset demonstrate that the generated knowledge base provides an excellent starting point for accelerating RCA tasks and improving network resilience.
\end{abstract}


\IEEEpeerreviewmaketitle

\section{Introduction}
\label{sec:introduction}
\IEEEPARstart{T}{he} reliability of communications networks is paramount in today's interconnected world.
However, the increasing complexity of these networks makes rapid and accurate root cause analysis (RCA) of outages a significant challenge.
Identifying the root causes of network issues often requires sifting through large volumes of heterogeneous data sources, such as support tickets, network logs, and performance metrics.
Traditional RCA methods rely heavily on manual analysis, which is not only time-consuming but also prone to human error, especially when dealing with large data sets.
Recent advances in artificial intelligence and machine learning (AI/ML) \cite{Canastro_2021, Wang_2024} have brought automation to the field, streamlining parts of the RCA process.
However, these methods often fall short when faced with the intricacies of human language and the contextual knowledge embedded in technical documents.
Capturing the nuances of written communication, especially in support tickets, remains a persistent bottleneck for existing AI/ML approaches.

The recent advancements in AI/ML have significantly impacted the field, Like with the introduction of large language models (LLMs), which represent a substantial innovation.
In the context of RCA, LLMs offer a transformative opportunity in this context.
As pre-trained models designed to understand and generate human-like text, LLMs are well-suited to process and synthesize unstructured textual data.
Thus, it can extract valuable information from support tickets, network logs, and technical documents, key sources of historical network knowledge.
These documents often contain insights gleaned from past incidents, including patterns, troubleshooting steps, and resolutions, which are vital for constructing a comprehensive knowledge base.
However, a significant challenge remains: effectively filtering this information while tailoring LLMs to grasp the specific nuances of RCA.
This requires designing task-specific prompts, fine-tuning LLMs on domain-specific data, and integrating them seamlessly into the RCA process.
Additionally, the development of a systematic workflow for deploying LLMs involves careful consideration of how they interact with existing systems, including data pre-processing, model inference, and the construction of a structured and actionable knowledge base.
By addressing these aspects, we can maximize the potential of LLMs to enhance the efficiency and accuracy of RCA.

Security and privacy are, therefore, critical considerations in the communication network industry, where vast amounts of sensitive data are processed daily \cite{penders2004privacy,Khan_2020}.
Consequently, the construction of a knowledge base must safeguard sensitive information by adhering to stringent security protocols and privacy regulations.
In this context, LLMs offer a distinct advantage, as they can be deployed locally, thereby minimizing data exposure while still delivering high performance.
Nevertheless, while inherent vulnerabilities in LLMs have been noted \cite{Yao_2024}, these risks can, in turn, be mitigated through robust implementation practices and careful model fine-tuning.

This paper presents, a novel approach leveraging LLMs to construct a knowledge base for automated RCA in communication networks.
In addition, we explore the capabilities of leveraging LLMs to extract and synthesize insights from support tickets, emphasizing the importance of integrating security and privacy considerations into our methodology.
Through critical examination of real resolved support tickets, we clarify the possibilities of each approach and its implications for enhancing the efficiency and accuracy of constructing a knowledge base in communication networks.
{\telcomodel} extracts valuable insights from these tickets and translates them into a structured knowledge base of association rules, enabling faster and more accurate diagnosis of network incidents.
Our key contributions are fourfold:

\begin{enumerate}
    \item We propose a framework, {\telcomodel}, that leverages LLMs to automate the construction of a RCA knowledge base using support ticket data.
    This framework is designed to address the specific challenges of contextual understanding and information synthesis in RCA tasks.
    \item We introduce a method for tailoring LLMs to the nuances of the task.
    This includes designing task-specific prompts, fine-tuning models on domain-specific data, and developing workflows that integrate LLMs seamlessly into the knowledge base construction process.
    \item We investigate three distinct approaches: fine-tuning LLMs, employing Retrieval-Augmented Generation (RAG), and a hybrid approach combining both.
    \item We evaluate these approaches on a real-world dataset of support tickets, demonstrating the effectiveness of the hybrid approach in generating accurate and contextually relevant RCA rules.
    Our analysis highlights the strengths and weaknesses of each approach and provides insights for practical deployment.
\end{enumerate}

The rest of this paper is organized as follows. 
An overview of related work is presented in Section \ref{sec:related_work}, followed by the 
problem statement and dataset description in Section \ref{sec:problem_statement_dataset}.
In Section \ref{sec:implementation}, we provide the details of our methodology and implementation.
In Section \ref{sec:results}, we present an analysis of our findings.
Finally, we draw the conclusions in Section \ref{sec:conclusion}.

\section{Related work}
\label{sec:related_work}

RCA is a crucial aspect of maintaining communications network reliability and ensuring sustained service delivery, particularly in the face of increasing network complexity and user demands.
Thus, it has attracted the attention of researchers, with its essential role in identifying the reasons for network interruptions and outages in an increasingly connected world \cite{Canastro_2021}, and in the upcoming 6G networks which will have to integrate existing networks \cite{Qiu_2023}.

Traditionally, the RCA process relies on information collected through a support system, where customers or end-users submit support tickets detailing the network issues they encounter.
Furthermore, analyzing root causes often requires sifting through diverse data sources like KPI time series, alarm events, and logs.
This process can be both time-consuming and expensive. Until recently, traditional RCA methods were limited to handling only one data type at a time and heavily relied on manual effort and expert knowledge \cite{wee2015method}.
Therefore, this often involved trial and error approaches, analysis of logs, network traffic and equipment configurations.
Additionally, automation of tasks such as data collection and basic diagnostics based on alarm matching existed, but complex analysis and decision-making were human-driven.
On the other hand, network monitoring systems often generate a high volume of alerts, making it difficult to distinguish root causes and side effects.
As a result, this could lead to extended troubleshooting times, under-performance and increased vulnerability to human error.
These aforementioned factors can collectively lead to reduced RCA effectiveness and ultimately a degradation of service assurance.

Using AI/ML to automatically point to the source of the problem has allowed engineers to troubleshoot faster and remove much of the guesswork when determining the cause of problems in large systems.
For instance, 
Sunita \textit{et al.} \cite{Yadwad2021ServiceOP} proposed a machine learning model trained on historical support tickets to predict future network failures.
Building upon this foundation, study \cite{szilagyi2012automatic} proposed a distinct approach centred around a scoring system.
This system automatically constructs a detection and diagnosis framework based on expert-labelled fault cases.
Then, it utilizes a scoring mechanism to assess the degree of similarity between a specific case and various diagnostic targets.
Similarly, Novaczki \cite{novaczki2013improved} proposed advancements in profiling and detection capabilities to enhance its overall effectiveness.
Another study \cite{Cana9590411} also uses machine learning to aid network automation to automate root cause analysis in cellular networks, enabling the quick diagnosis and tracking of anomalies by correlating log files from negative-impacting events across network levels using defined rules and a graph-based approach to dependencies.

However, these approaches use classical machine learning algorithms, e.g., ridge regression, time series, linear models, etc., that require significant manual effort in feature engineering and may not capture the intricate relationships between symptoms, root causes and solutions.
Furthermore, a significant challenge associated with traditional RCA methodologies lies in their limitations when managing synonyms, understanding semantic nuances, and data context.

The second generation of machine learning algorithms for RCA is divided into three broad classes: 
Trace-based, e.g., TraceAnomaly \cite{Liu_2020} that was proposed for microservice systems, 
Rule-based, with, e.g., the mining of association rules \cite{Phan_2021}, and
Graph-based, with, e.g., the diagnosis of the root causes of incidents in a large industrial system \cite{Wang_2021}.
These approaches definitively improved over the first generation, but are still lacking the automated understanding of, e.g., the log tickets.
Several recent studies have explored the application of LLM to RCA tasks in network environments.
In \cite{Ahmed_2023}, Ahmed \textit{et al.} propose a method that fine-tunes a Generative Pre-trained Transformer (GPT) model.
Their approach allows the model to generate text, transforming incident summaries directly into root cause identifications and corresponding suggestions for mitigation strategies.
With a similar method, Jin \textit{et al}. \cite{jin2023assess} present a work that proposes fine-tuning a GPT model using domain-specific data.
This fine-tuned model then predicts outage summaries, facilitating a deeper understanding of root causes and guiding post-hoc maintenance efforts.
While previous research has demonstrated the effectiveness of fine-tuned LLMs in generating root cause identifications and fault summaries, there is an exciting opportunity to explore how augmenting these models with external knowledge sources can further improve their accuracy and completeness in identifying and mitigating network issues.
This line of inquiry stems from the inherent limitations of fine-tuned LLMs. While these models are good at text generation, they may lack access to the broader context and domain-specific knowledge crucial for pinpointing the root causes of complex network anomalies.
By incorporating external knowledge sources, such as historical sensitive support tickets on the customer side or technical documentation, we can potentially equip these models with a richer understanding of network behaviour and relevant troubleshooting procedures.

One potential solution to address the aforementioned limitations of fine-tuned LLMs lies in the application of a few-shot Retrieval-Augmented Generation \cite{lewis2020retrieval}.
This approach aims to bridge the gap in knowledge by strategically incorporating external knowledge sources into the LLM's reasoning process.
Within the domain of RCA, several studies have emerged that explore the application of RAG concepts (e.g., \cite{chen2024automatic, zhang2023pace}).
These studies leverage publicly available service LLMs API, such as ChatGPT, to implement the RAG framework and inferences based on their data.
However, this approach raises concerns regarding data privacy, a critical principle within the communication network industry.
The RAG concept offers a promising avenue to address the limitations of fine-tuned LLMs, this approach holds significant potential for enhancing the accuracy and comprehensiveness of RCA tasks. However, further research is necessary to explore the practical implications and considerations associated with its implementation for highly confidential pieces of information and employ them for building a knowledge base.
While the prospect of leveraging LLMs for direct data access and network anomaly analysis is enticing, several critical challenges impede its practical application \cite{assert-llms}.
One of them is the security concerns, integrating LLMs directly into customer-side operations introduces significant security vulnerabilities.
LLMs, with their extensive access requirements, could become potential entry points for malicious actors, putting a risk to sensitive customer data and network integrity.
In addition, reliability and trust are also concerns as the inherent nature of LLMs raises concerns regarding the reliability and trustworthiness of their generated analyses.
Their probabilistic and black-box nature makes it difficult to ascertain the validity and provenance of their conclusions, potentially leading to misinterpretations and erroneous actions.
On the other hand, knowledge bases play a crucial role in root cause analysis by providing structured information, historical data, and domain-specific knowledge that are essential for conducting a thorough investigation.

In contrast to the aforementioned research, our proposal takes an approach to leveraging the capabilities of LLMs for RCA within communication networks.
While previous studies have focused on integrating external knowledge sources or fine-tuning LLMs to directly generate root cause identifications and outage summaries, our approach aims to utilize LLMs as a key tool in constructing a comprehensive knowledge base for RCA tasks, drawing upon data extracted from support tickets.
Moreover, by utilizing support ticket data for knowledge base construction, we inherently address concerns related to data privacy and confidentiality.
Furthermore, we investigate the applicability of LLMs for on-premise deployment through three distinct approaches: fine-tuned LLMs, RAG-enhanced LLMs, and a hybrid approach that combines both methodologies.

\section{Problem statement \& dataset}
\label{sec:problem_statement_dataset}

In this section, we describe the main problem we want to solve and present the actual dataset that will serve as the basis for our investigation.


\subsection{Problem Statement}

We assume we are given a collection of historical network support tickets $
\mathcal{D} = \{(x_i, y_i)\}_{i=1}^N $ containing information about network anomalies experienced by customers.
Where $x_i$ is the textual description of a network anomaly (e.g., logs, customer complaints, product impacts), and $y_i$ is the corresponding root cause analysis and solution.
The main objective is to develop an automated system that analyzes these network tickets and extracts information which is then translated into an RCA knowledge base ($\mathcal{K}$) structured in association rules format.
This RCA knowledge base will then be used to diagnose and resolve future network incidents more quickly and efficiently, as well as improve automation service assurance.
Formally, we aim to learn a function $f: \mathcal{X} \rightarrow \mathcal{Y}$, where $\mathcal{X}$ is the space of anomaly descriptions, and $\mathcal{Y}$ is the space of RCA rules.

\subsection{Dataset}

Effective implementation of an RCA application utilizing LLMs hinges on a well-curated and pre-processed dataset.
This section delves into the data collection and preparation strategies employed in this research.
Our study is based on a use case from a network operations company, which leverages a dataset of distinct support tickets collected from an internal support portal over several years.
These tickets represent resolved network incidents that have been reported by customers and have been resolved by technical experts.
Each contains free-text descriptions of network anomalies, potentially including logs, expert analyses, and proposed solutions.

\begin{table}[hbt]
    \centering
    \begin{tabular}{|L{1.2cm}|L{6.4cm}|}
        \hline
        \textbf{Attribute} & \textbf{Description} \\
        \hline
        Title  & The general title of the ticket, e.g., UE increased latency during the afternoon. \\
        \hline
        
        Product & The product(s) impacted by the incident, e.g., Radio Dot. \\
        \hline

        Anomaly & A detailed of the issue faced by the customer, including logs, symptoms, and any relevant information. E.g., the customer reported intermittent connectivity issues affecting users between 3:00 PM and 4:00 PM. Logs from the core router show a series of packet loss events and increased latency during this time. The command shows log messages | including "packet loss" revealing frequent drops, coinciding with the reported time window. Additionally, end users experienced slow page load times and difficulty accessing internal applications. Symptoms include delays in voice and video calls and a significant drop in throughput as shown by the command show interface statistics during the incident.
 \\
        \hline

        Technical analysis & Technical analysis and troubleshooting notes regarding the cause of the incident in the network. For instance, investigated increased latency using the command show log messages | include latency to collect relevant logs. Identified packet drops between 3-4 PM due to high CPU utilization on the core router. \\
        \hline
        Propose solution & A step-by-step resolution provided by a technical expert to address the network incident, including commands, monitoring steps, and expected behaviour after implementation.
        For example, apply a temporary QoS policy using set \textit{policy-options policy-statement HIGH-PRIO term 1 }then priority high. \\

        \hline
    \end{tabular}
    \caption{Ticket attributes}
    \label{tab:ticket_attr}
\end{table}

In our pipeline, we expect each ticket to have the attributes outlined in Table~\ref{tab:ticket_attr}, with content in a free-text format, including logs and network configurations.
These attributes are crucial for building the RCA knowledge base, as they provide a comprehensive understanding of a true network anomaly and its resolution in real-world scenarios.
Consequently, the data can be leveraged to infer potential future issues within the network or be integrated into an automated RCA system.

\section{Methodology}
\label{sec:implementation}

This section outlines the implementation of our methodology, {\telcomodel}, designed to automatically build a knowledge base for RCA applications using LLMs. Within {\telcomodel}, we explore three distinct approaches: fine-tuning, Retrieval-Augmented Generation (RAG), and a hybrid method that combines the strengths of both fine-tuning and RAG concepts.


\subsection{Designing prompts for RCA application}

To train or fine-tune the LLM, this data is first divided into smaller pieces called tokens.
Once divided into tokens, the LLM can begin to learn the building blocks of language from these smaller units.
The size of the tokens can have a significant impact on the performance of the LLM \cite{xue2024repeat}.
In the RCA domain, it is important to identify the root cause of a network anomaly from a large amount of text data.
However, using a huge number of tokens can be challenging as the performance of an LLM can be reduced, potentially hindering its ability to learn complex relationships between words.
For instance, LLAMA 2-7B \cite{touvron2023llama} performs best when processing sequences of up to 4,096 tokens. This means the input text, broken down into its basic units (tokens), should ideally be no more than 4,096 tokens in length.
\begin{figure}[ht]
    \centering
    \includegraphics[width=1\columnwidth]{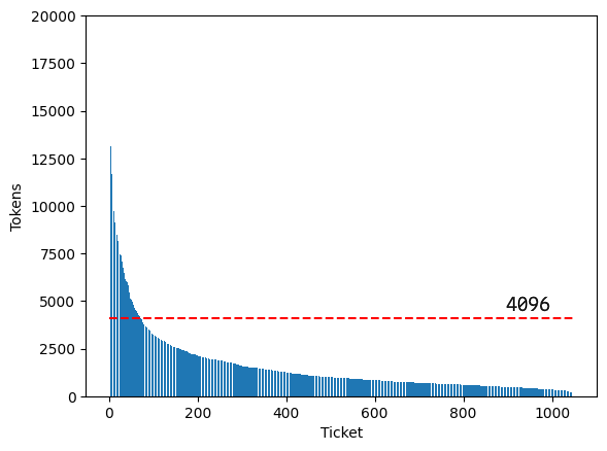}
    \caption{Distribution of token lengths in the dataset.}
    \label{fig:token-statistic}
\end{figure}

\begin{algorithm}[ht]
\caption{Prompt-Aware Text Chunking.}
\label{alg:split_chunks}
\begin{algorithmic}[1]
\Require Textual data $\mathcal{D}$, pre-defined prompt $\mathcal{P}$, max\_token
\Ensure Set of chunks $C$

\State Tokenize prompt: $p \gets \text{Tokenize}(\mathcal{P})$  
\State Tokenize data: $d \gets \text{Tokenize}(\mathcal{D})$  
\State Calculate remaining tokens: $r \gets \text{max\_token} - \text{len}(p)$  
\State Compute max chunks: $\text{max\_chunks} \gets \left\lceil \frac{\text{len}(d)}{r} \right\rceil$  
\State Initialize chunks: $C \gets \emptyset$  

\For{$i \in [0, \text{max\_chunks} - 1]$}  
    \If{$i \neq \text{max\_chunks} - 1$}  
        \State chunk $\gets p \cup d[i \cdot r : (i + 1) \cdot r]$  
    \Else  
        \State chunk $\gets p \cup d[i \cdot r :]$ \Comment{Last chunk takes remaining tokens}  
    \EndIf  
    \State $C \gets C \cup \{\text{chunk}\}$  
\EndFor  

\State \Return $C$
\end{algorithmic}
\end{algorithm}

As shown in Figure \ref{fig:token-statistic}, a part of our dataset contains entries exceeding the maximum token limit (4,096 tokens) for some LLMs. 
This means some pieces of text are too long for the model to process effectively.
To address the above challenge, we designed Algorithm \ref{alg:split_chunks} which processes our textual data and divides it into suitable pieces (chunks) along with our predefined prompts.
This process effectively transforms the text into a format that the LLM can understand and analyze for further processing.
For more details on the impacts of the tokenization process, please refer to \cite{Sachidananda2021}.

In addition, to guide the LLM in processing our data, a set of pre-defined prompts with specific formats was constructed:
\begin{itemize}
    \item[1.] \textit{Network Anomaly Analysis Prompt}: this prompt instructs the LLM to identify anomalies within network data and their corresponding symptoms. The prompt is formatted as: \begin{quote}
        \textit{Identify and extract network anomaly and its symptoms into a pair of insights, logs in products [product impacts]: [textual anomaly description in ticket].}
    \end{quote}
    
    \item[2.]  \textit{Root Cause and Solution Extraction Prompt}: this prompt guides the LLM to extract potential root causes impacting specific products and their corresponding solutions. Hence, our prompt is formatted as:
    \begin{quote}
        \textit{Analyze and extract possible pair of root cause and solution. Here are the references for: [enter character] root cause analysis: [textual of root cause]. [enter character] solution: [textual of solution in ticket].}
    \end{quote}
    
    \item[3.] \textit{Combine all results}: this prompt serves as a summarization step, whenever our data is split into multiple chunks during the preceding steps.
    This prompt instructs the LLM to leverage the results from the two previous prompts by incorporating "[result from first prompt]" for the anomaly and "[result from second prompt]" for the identified root cause and solution.
    Thus, this prompt is formatted as:
    \begin{quote}
        \textit{Given the network anomaly analysis: ["Results from the first prompt"]; root cause and solution analysis: ["Result from the second prompt"]. Your task is to conduct all possible association rules and format as: [network anomaly, product impact, root cause, solution]. Each rule will be on one line.}
    \end{quote}
    
\end{itemize}

By utilizing these pre-defined prompts, {\telcomodel} effectively guides the LLM through the analysis process, providing clear instructions on the required tasks and the appropriate formats for presenting results.
This strategic approach not only enhances the efficiency and consistency of the analysis but also enables the LLM to generate actionable insights and recommendations tailored to specific network anomalies and their root causes.
Furthermore, by designing prompts to align with desired outcomes and incorporating references to contextual information, we simplify the LLM’s comprehension of RCA-specific terminology and the complexities involved in analyzing and resolving network anomalies.
Consequently, the LLM leverages this structured guidance to produce more accurate and contextually relevant analyses, which, in turn, support improved decision-making and more effective problem-solving in network management and optimization efforts.

\subsection{Fine-tune LLM for RCA}

Training a LLM from scratch involves designing its architecture and learning parameters using a massive dataset, often requiring billions of tokens.
This process demands substantial computational resources, time, and expertise in curating high-quality, domain-specific data.
In contrast, fine-tuning leverages pre-trained LLMs, adapting them to specific tasks or domains with smaller, targeted datasets.
Fine-tuning is a best practice in intelligent systems development, as pre-trained models already possess general linguistic and semantic knowledge, eliminating the need to relearn foundational concepts.
This makes fine-tuning computationally efficient and cost-effective, especially for narrow applications where large datasets are unavailable.
We utilize pre-trained LLMs like LLAMA3 and Gemma as base models for our evaluation.
These models are selected for their strong performance in natural language understanding and generation.
Our goal is to adapt them for RCA tasks within the company's products and network, enhancing their ability to identify issues and generate actionable solutions.

Let \(\mathcal{M}\) be a pre-trained LLM (e.g., LLaMA, Gemma) with parameters \(\theta\). Fine-tuning involves adapting \(\mathcal{M}\) to the RCA task by minimizing a loss function \(\mathcal{L}\) over the dataset \(\mathcal{D}\):
\[
\mathcal{L}(\theta) = \sum_{i=1}^{N} \ell(f_{\theta}(x_{i}), y_{i}),
\]
where \(\ell\) is a task-specific loss function (mean squared error in our case). Fine-tuning is performed using Low-Rank Adaptation (LoRA) \cite{hu2022lora}, which reduces the number of trainable parameters by decomposing the weight matrices \(W \in \mathbb{R}^{m \times n}\) into low-rank components:
\[
W = W_{0} + \Delta W, \quad \Delta W = A \cdot B^{T},
\]
where \(A \in \mathbb{R}^{m \times r}\) and \(B \in \mathbb{R}^{n \times r}\) are low-rank matrices, and \(r \ll \min(m, n)\). This approach significantly reduces computational costs while maintaining model performance.

\begin{figure}[ht]
    \centering
    \includegraphics[width=0.99\columnwidth]{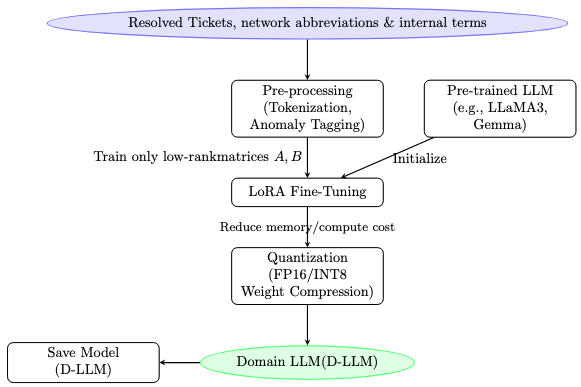}
    \caption{Fine-tuning phase with LoRA and quantization for Domain LLM (D-LLM).}
    \label{fig:fine-tune-domain-llm}
\end{figure}

Figure \ref{fig:fine-tune-domain-llm} illustrates the process of fine-tuning the LLM within our {\telcomodel} framework.
The fine-tuning process involves training the LLM on a curated dataset of support tickets, formatted as input-output pairs.
It enables the effective transfer of knowledge obtained from pre-existing data sources, thereby augmenting the comprehension capabilities of these models \cite{chen-fine-tune-llm9377810}.
The input consists of the anomaly description and associated symptoms, while the output is the corresponding root cause analysis and proposed solution.
This aims to transfer knowledge specific from support tickets into the LLM, enhancing its ability to understand and analyze similar future issues.
The primary approach employed in this step involves the implementation of supervised fine-tuning with instruction (SFTI) in conjunction with LoRA.
This approach is well-known and widely used for fine-tuning LLMs.
However, due to limitations in GPU capacity, we also apply the quantization technique \cite{lin2024awq}, which reduces the model to 16 bits, ensuring more efficient resource usage. 
Additionally, the maximum sequence length is set to 4906, as some of the data in our dataset includes logs.
This extended sequence length is particularly beneficial, as it helps the model better understand anomalies and, as a result, adjust its weights more effectively.
In terms of hyper-parameters, the number of epochs for fine-tuning is set to 4.
This decision was made after careful testing across a range of epochs from 3 to 10, with a focus on minimizing overfitting and maximizing performance.
For further references on selecting the optimal number of epochs, see \cite{xia2024understanding}.
Furthermore, the batch size is set to 4, as we are constrained by GPU capacity.
This smaller batch size ensures that the model can still process the data efficiently within the available hardware limitations, without sacrificing too much on the fine-tuning performance.
Following the fine-tuning process, the resulting model, a Domain LLM (D-LLM), will possess enhanced comprehension of network anomalies. 
This D-LLM leverages the acquired knowledge from support tickets to analyze root causes and propose solutions for future network anomalies.

In addition to the dataset described in the previous section, fine-tuning the LLM on network abbreviations, both internal product abbreviations specific to the company and those commonly used in RCA domains, represents a critical step in enhancing its understanding.
Thus, a dedicated set of abbreviations is leveraged as part of the fine-tuning process.
Abbreviations, such as QoS (Quality of Service), serve as concise representations of complex concepts, entities, or processes, enabling streamlined communication and more efficient information exchange.
These abbreviations will also be incorporated into the dataset as paired examples.
Incorporating these abbreviations into the fine-tuning process not only enhances the LLM’s language comprehension but also aligns its knowledge with the specialized linguistic conventions and terminologies prevalent in the company’s network environment.
This alignment ensures that the LLM is better equipped to interpret and analyze domain-specific content accurately.

\subsection{Employ Retrieval-Augmented Generation concept}

The RAG paradigm \cite{Ahn-rag-9982598} has emerged as a powerful framework for enhancing the capabilities of LLMs in generating contextually relevant and informative text.
This approach strategically combines elements of retrieval-based methods, which leverage pre-existing knowledge sources, with generative models capable of producing novel and coherent text.
Employing the RAG concept in the domain of network RCA presents exciting opportunities to streamline and enrich the process of identifying and addressing network issues.

At its core, RAG operates on the principle of integrating external knowledge sources, such as databases, documents, or support tickets in our case, into the generation process of LLMs.
This integration of retrieved information with generative text creation enables models to draw upon a vast pool of pre-existing knowledge while still retaining the creativity and flexibility inherent in generative approaches.
In the context of constructing a knowledge base for network anomaly in RCA, RAG offers several key benefits:

\begin{itemize}
    \item \textit{Enhanced knowledge integration}: by combining data from resolved support tickets, network documentation, or historical RCA reports into the generation process, LLMs can access a rich repository of domain-specific knowledge.
    This knowledge serves as a foundation upon which the model can generate insightful root cause analyses and proposed solutions tailored to specific network anomalies.

    \item \textit{Improved contextual understanding}: RAG allows LLMs to contextualize their outputs based on the retrieved information.
    When tasked with generating root cause analyses for network anomalies, models can retrieve relevant historical data (e.g., past incidents with similar symptoms) and incorporate it into their generation process, ensuring that previous experiences and observations inform the generated analyses.
    This makes the output more reliable and more accurate.

    \item \textit{Iterative refinement}: RAG supports iterative refinement of generated outputs based on feedback from the retrieved knowledge sources.
    As models generate root cause analyses and proposed solutions, they can dynamically adjust their outputs based on the relevance and reliability of the retrieved information, iteratively improving the quality and accuracy of their response.
\end{itemize}

\begin{figure}[ht]
    \centering
    \includegraphics[width=1\columnwidth]{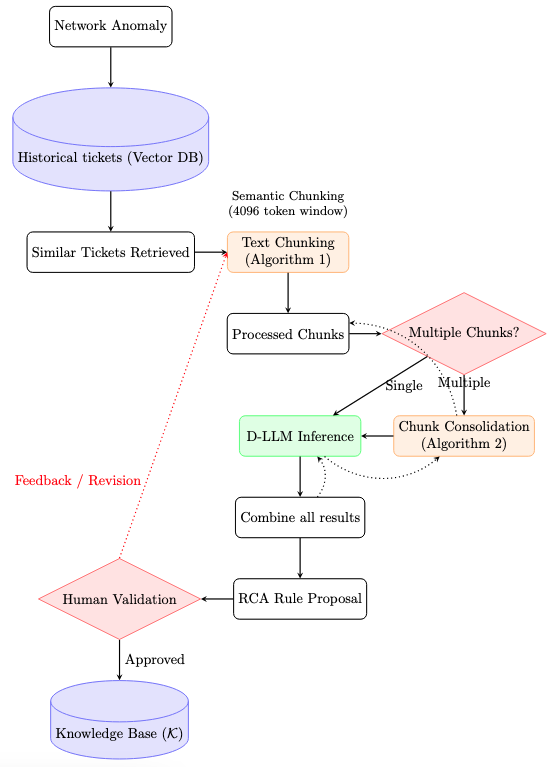}
    \caption{Inference pipeline for automated RCA rule generation in {\telcomodel}. The system combines retrieval-augmented generation with human validation feedback (optional) loops.}
    \label{fig:rag-llm-inference}
\end{figure}

Building upon the preceding approach, {\telcomodel} integrates external knowledge sources (e.g., historical of resolved support tickets, internal documents) into the LLM’s reasoning process.
Let $K = \{(a_j, r_j, s_j)\}_{j=1}^M$ be a knowledge base of historical anomalies $ a_j $, root causes $ r_j $, and solutions $ s_j $.
Given a new anomaly $ x $, {\telcomodel} retrieves the most relevant entries from $ K $ using a similarity function $\text{sim}(x, a_j)$:
\[
\text{sim}(x, a_j) = \frac{\phi(x)^T \phi(a_j)}{\|\phi(x)\| \cdot \|\phi(a_j)\|},
\]
where \( \phi \) is an embedding function (e.g., BERT or Word2Vec). The retrieved entries are then used to augment the input to the LLM, enabling it to generate more contextually relevant RCA rules.
As illustrated in Figure \ref{fig:rag-llm-inference}, when given an input network anomaly, the vector database is queried to retrieve the most relevant historical network anomalies, along with their associated symptoms and the corresponding expert-identified root cause analyses and proposed solutions, which contains in tickets, based on a similarity threshold, e.g. 70\%.
%

%
\begin{algorithm}
\caption{Consolidate chunks}\label{alg:consolidation_chunks}
\begin{algorithmic}[1]
\Require Processed chunks $C = \{c_1, ..., c_n\}$
\Ensure Consolidated context $S$
\If{$|C| > 1$}
    \State Initialize $S \gets \emptyset$
    \For{each chunk $c_i \in C$}
        \State Analyze $c_i$ for key information by D-LLM
        \State $S \gets S \cup \text{ExtractContext}(c_i)$
    \EndFor
    \State Generate consolidated report
\Else
    \State Direct processing of $c_1$
\EndIf
\State \Return $S$
\end{algorithmic}
\end{algorithm}

Subsequently, the retrieved information (including relevant historical network anomalies, their symptoms, and corresponding root cause analyses/solutions) is combined with pre-defined prompts.
This combined data is then transferred to Algorithm~\ref{alg:split_chunks} for segmentation into smaller chunks.
This chunking process optimizes the performance of the D-LLM during analysis by ensuring manageable input sizes and focused context processing.
Following the chunking process, if more than one chunk exists, Algorithm~\ref{alg:consolidation_chunks} is employed to ensure that all information is captured and analyzed by the D-LLM before combining the results.
After generation, the formulated RCA rules will undergo a careful evaluation process conducted by our technical specialists.
This evaluation ensures the legitimacy and alignment of the rules with both our internal systems and those deployed within our customer base prior to integration into production environments.

\subsection{Performance metrics}
\label{sec:performanceM}
The evaluation in the next section employed various metrics encompassing lexical and semantic similarity, as is common in LLM evaluation.
In detail, lexical similarity was evaluated using Cosine similarity, BLEU, METEOR, and ROUGE scores, while semantic similarity was assessed using the BERT score.
This varied evaluation strategy offers a comprehensive assessment of the quality and faithfulness of the knowledge bases constructed through each technique.

\begin{figure}
    \centering
    \includegraphics[width=1\columnwidth]{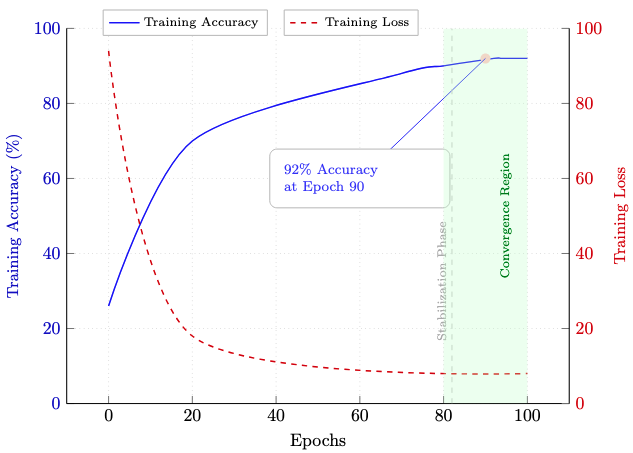}
    \caption{Results on the training of the Word2Vec model for the cosine similarity metric.}
    \label{fig:train_word2vec}
\end{figure}

BLEU \cite{papineni2002bleu} is a widely used metric for evaluating the lexical similarity between machine-generated text and human-written references.
It functions by calculating the n-gram overlap, which essentially measures the degree to which sequences of words (of length n) appear in both the generated text and the reference text. This approach emphasizes precision, ensuring that the generated text contains content found in the reference.
Similarly, METEOR \cite{banerjee2005meteor} incorporates both precision and recall, employing more intricate text processing and scoring mechanisms.
ROUGE \cite{lin2004rouge} is a prevalent metric for evaluating text summarization and emphasizes recall.
In contrast, BERT score \cite{zhang2019bertscore} leverages pre-trained BERT models to gauge semantic similarity.
To assess the cosine similarity between texts, we built a word vector model using 80\% of our ticket dataset.
The word vector is based on the Word2Vec\protect\footnotemark[1] \footnotetext[1]{https://code.google.com/archive/p/word2vec/} developed by Google.
This helps to learn word representations, also known as word embeddings, by analyzing the context in which words appear within a large corpus of text.
These word embeddings capture semantic relationships between words, allowing us to quantify the similarity between texts based on the similarity of their word embeddings.
As shown in Fig. \ref{fig:train_word2vec}, our model achieved a 92\% accuracy rate on a held-out test set, comprising 20\% of the tickets, after around 90 epochs of training.
The vocabulary size used during training encompassed 12,008 unique words focused on the RCA domain.

\section{Results and Findings}
\label{sec:results}

This section presents the key outcomes of the proposed methodology for automated knowledge base construction in RCA, leveraging fine-tuned LLMs, RAG, and hybrid approaches.
The evaluation is divided into two subsections. 
The first subsection focuses on the impact of prompts on the methodology, while the second subsection assesses the performance of the methodology in building a knowledge base for RCA tasks.
\subsection{Experiment dataset}

To evaluate {\telcomodel}, we employ a real-world incident ticket dataset consisting of 1,049 support cases collected over several years from the internal support portal.
Each ticket includes detailed free-text anomaly descriptions, log excerpts, expert analyses, and solutions.
To illustrate the ticket, Table \ref{tab:example_sp_ticket} provides an anonymized sample support ticket. 
This example highlights the structure and content of the data used in our evaluation and demonstrates the type of real-world cases on which our approach is applied.

\begin{table}[hbt]
    \centering
    \begin{tabular}{ | L{1.8cm} | L{5.8cm} | }
        \hline
        \textbf{Attribute} & \textbf{Information} \\
        \hline
        
        ID & AA2854542 \\
        
        Title & Unable to configure DF bit for data transfer \\
        
        Products & Product A, Product B \\
        
        Issue Category & Configuration/Provisioning. \\
        
        \hline
        
        Anomaly & We are encountering an issue where we cannot enable the DF bit for data transfer between Product A and Product B. This functionality was previously available but seems to be missing in the current version. This may cause our system down after the upgrade. \\
        
        \hline
        
        Root Cause Analysis (RCA) & We have made a diagnosis and there are 2 possible problems:
        
        \textbf{Software Version}: The issue might be related to the software version currently running on Application B (version 2.0). The "clear-df" option might only be supported in versions 2.1 and above.
        
        \textbf{Missing Configuration Option}: Our analysis of the configuration options for data transfer suggests that the "clear-df" option is not explicitly available. This could be due to limitations in the current configuration tools. \\ 
        \hline

        Solution & \textbf{Upgrade Application B}: We recommend upgrading Application B to version 2.1 or later. This version is confirmed to support the "clear-df" option.
        
        \textbf{Define Tunnel with Specific Key}: As an alternative solution, you can define a GRE tunnel with a specific key for the data transfer between Application A and B. This type of tunnel allows configuring the "clear-df" option.
        
        \textbf{Additional Information}:
        The "clear-df" option ensures that data packets are not fragmented during transmission, which can improve performance in certain scenarios.
        
        \textbf{Reference ticket}: ABC19666 provides details on configuring the "clear-df" option with a specific GRE key.
        
        \textbf{Next Steps}:
        Please let us know if you prefer to upgrade Application B or proceed with defining a GRE tunnel with a specific key. We can assist you further with the chosen solution.\\
        
        \hline
    \end{tabular}
    \caption{Support ticket example\protect\footnotemark[2].}
    \label{tab:example_sp_ticket}
\end{table}
\footnotetext[2]{All sensitive data were anonymized and the context was simplified.} 

\begin{figure}[ht]
    \centering
    \includegraphics[width=1\columnwidth]{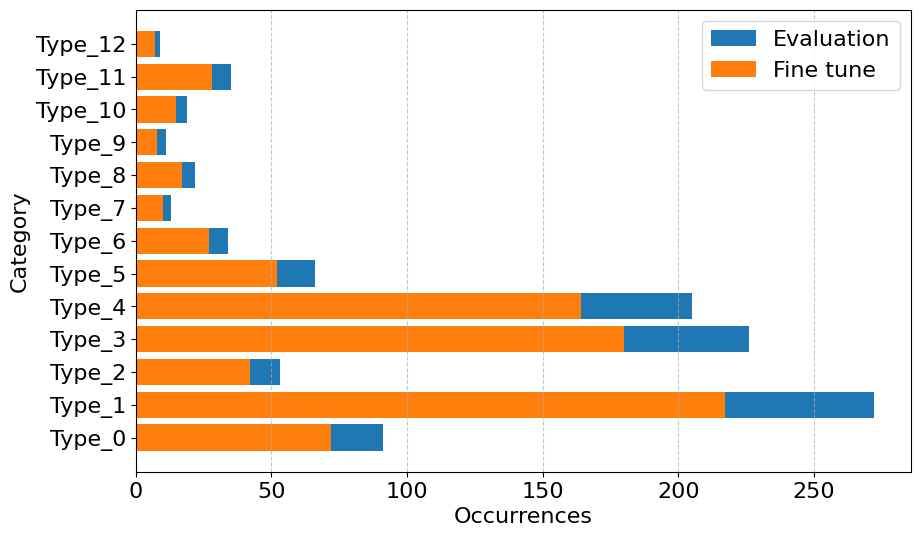}
    \caption{Network issue distribution: Fine-tune vs. evaluation sets \protect\footnotemark[3].}
    \label{fig:data-train-eval}
\end{figure}

The dataset covers 13 distinct network issue classes, reflecting the operational diversity typically encountered in production network environments.
These classes span a wide range of problem types, including traffic anomalies, software errors, hardware failures, and configuration issues.
As shown in Figure \ref{fig:data-train-eval}, the distribution of these anomaly types within the dataset reveals an imbalance, with certain types having significantly more instances than others.
This disparity reflects the real-world occurrence patterns of network issues, where some problems are inherently more frequent than others.
Moreover, this diversity in network anomaly types is crucial for fine-tuning the model.
By exposing the LLM to a wide range of anomalies, it becomes better equipped to handle various network issues, extract relevant information across different root causes, and propose appropriate solutions.
To illustrate, imagine the LLM as an automated detective; the broader the set of cases it has investigated (i.e., been fine-tuned on), the better it becomes at identifying patterns and solving future network anomalies effectively.
The utilization of this methodology offers two primary benefits.
First, it enhances the capacity of the model for generalization, enabling effective application of learned principles to new, unseen data.
Second, it improves model robustness in real-world scenarios, ensuring reliable performance under varied conditions.

\footnotetext[3]{Category names were de-identified to ensure anonymity.}

\begin{table}[ht]
    \centering
    \begin{tabular}{ | L{1.8cm} | L{5.8cm} | }
        \hline
        \textbf{Attribute} & \textbf{Description} \\
        \hline
        
        id      & Represents a unique identifier for a ticket. \\
        title   & Short title to describe the network anomaly. \\
        anomaly & Textual describes unusual data/behaviour within customer network reports, potentially indicating network malfunctions. \\
        root cause & Textual describes root cause analysis from technical experts for the anomaly above. \\
        solution & Textual describes the proposed solution for the above root cause from technical experts. \\
        issue category & Network category for the above root cause. \\
        products & List of products impacted by the above anomaly and root cause. \\
        \hline
    \end{tabular}
    \caption{Attributes in a support ticket.}
    \label{tab:description_input_ticket}
\end{table}

However, simply putting raw data into an LLM is not enough in practice.
To effectively fine-tune the model, a meticulous data pre-processing stage is essential.
This stage involves constructing a fine-tuning dataset that serves as the foundation for adapting the LLM.
The primary dataset consists of three key components:
\begin{itemize}
    \item \textit{Resolved support tickets:} The core foundation is a collection of resolved support tickets representing various network issues.
    These tickets serve as real-world case studies for the LLM to learn from.
    The dataset contains diverse textual data, and a crucial step involved transforming the resolved support tickets into a structured format, as summarized in Table \ref{tab:description_input_ticket}.
    
    \item \textit{LLM prompts:} these prompts act as guiding principles for the LLM, instructing it on how to interpret the input data (e.g., network anomaly description) and the desired format of the output (e.g., root cause analysis). The details of the design prompt will be discussed in the next section.
    
    \item \textit{Example output data:} to facilitate supervised learning, the fine-tuning dataset incorporates corresponding examples of output data (e.g., root causes, solutions) alongside the input data (e.g., network anomaly description).
    These example outputs represent the desired outcome for the LLM, such as accurately identifying the root cause of the network anomaly.
    This inclusion operates similarly to supervised learning paradigms, where labelled examples offer the model a reference for understanding the correct adjustment of weights between input and desired output during the fine-tuning process.
\end{itemize}

\begin{table}[ht]
    \centering
    \begin{tabular}{ | L{1.8cm} | L{5.8cm} | }
        \hline
        \textbf{Attribute} & \textbf{Information} \\
        \hline
        
        anomaly & Unable to configure the DF bit for data transfer \\
        
        \hline
        Root cause 1 & Software version incompatible. \\
        Products impact & Product B \\
        Solution & Update software on Product B to v2.1 and above. \\
        \hline
        Root cause 2 & The configuration tools do not offer the "clear-df" option for the chosen data transfer method. \\
        Products impact & Product A, Product B. \\
        Solution & Consider defining a GRE tunnel with a specific key for the data transfer, as this method typically allows configuring the "clear-df" option.\\
        
        \hline
    \end{tabular}
    \caption{An example of RCA rule derived from the support ticket example in Table \ref{tab:example_sp_ticket} \protect\footnotemark[4]. }
    \label{tab:example_RCA_rule}
\end{table}
\footnotetext[4]{All of the information was anonymized.}

Finally, a critical step in the process involves constructing a dataset specifically tailored for fine-tuning and evaluating the LLM.
This process requires systematically collecting relevant textual data from support tickets and incorporating carefully defined prompts.
These prompts play a dual role: they guide the LLM during its analysis and establish the desired output format, ensuring alignment with the task's objectives.
As a result, the dataset becomes the cornerstone for both fine-tuning and evaluation, enabling the LLM to perform root cause analysis tasks with precision and consistency.
Building upon the techniques outlined earlier, we curated a dataset comprising 3,147 examples of paired inputs and outputs.
These examples are derived from resolved tickets, and serve as a structured foundation for training and evaluating the model.
For instance, in one example pair, the expected output from a ticket (e.g., Table~\ref{tab:example_sp_ticket}) is an RCA rule, as illustrated in Table~\ref{tab:example_RCA_rule}.

Following the implementation of the pre-processing procedures described above, the dataset was divided into two distinct components: 80\% of the data, encompassing all network anomaly types, was allocated for fine-tuning the LLMs, while the remaining 20\% was reserved for evaluating the efficacy of the models in the downstream task of constructing a knowledge base for RCA.
This strategy ensures that the model is exposed to a diverse array of patterns and scenarios within the data, thereby facilitating effective learning and improving its generalization capabilities.
The ratio 80/20 split is widely recognized in machine learning research, especially in fine-tuning LLM, as a standard practice for partitioning datasets due to its balanced approach to fine-tuning and testing \cite{goodfellow2016deep}.
Indeed, allocating too small a portion for evaluation may fail to capture the variability of unseen data, while reserving a larger fraction for evaluation could limit the model’s learning potential by reducing the amount of fine-tune data available.
By adopting the 80/20 split, we achieve a practical balance that aligns with established best practices in the field, ensuring both robust fine-tune and meaningful evaluation.


\subsection{Impact of prompts}

\begin{table}[ht]
    \centering
    \resizebox{\columnwidth}{!}{%
    \begin{tabular}{lccccc} 
        \hline
        \textbf{Model}      & \textbf{Cosine}    & \multirow{2}{*}{\textbf{BLEU}}   & \multirow{2}{*}{\textbf{ROUGE}} 
                                                 & \multirow{2}{*}{\textbf{METEOR}} & \multirow{2}{*}{\textbf{BERT}} \\
        \textbf{LLAMA3-8B} & \textbf{similarity} &  &  &  &  \\
        \hline
        Without prompt & 0.208 & 0.032 & 0.060 & 0.059 & 0.811 \\
        With prompt & 0.589 & 0.118 & 0.162 & 0.161 & 0.866 \\
        \hline
    \end{tabular}
    }
    \caption{Impact of prompts on performance.}
    \label{tab:impact_of_prompts}
\end{table}
To investigate the impact of prompts on LLMs, we investigated the influence of prompts on LLMs by evaluating the performance of LLAMA3-8B, a state-of-the-art LLM at the time of our evaluation.
We compared the performance of the models with and without prompts using a variety of evaluation metrics, as described in the previous section.
The results, demonstrated in Table \ref{tab:impact_of_prompts}, show that LLM performs significantly better at all metrics when given prompts.
For example, the cosine similarity score increases from 0.208 to 0.589, the BLEU score rises from 0.032 to 0.118, and the ROUGE score increases from 0.060 to 0.162.
This suggests that prompts can help LLMs to understand the task at hand better and to generate more relevant and informative outputs.
In addition, it shows that prompts can help LLMs to focus on the most relevant information in the input information.
When provided with a prompt, an LLM can use it to identify the key elements of the task and focus its attention on those elements.
Therefore, this can help the LLM to avoid getting sidetracked by irrelevant information.

Throughout the experiment, we found that prompts can help LLMs to learn task-specific vocabulary, i.e., tasks in RCA domain in our context.
Even though LLMs are trained on a massive amount of text data, this data may not include all of the vocabulary that is relevant to a particular task.
Therefore, prompts can help LLMs learn task-specific vocabulary by providing them with guides/examples of how the vocabulary is used. This can help the LLM to generate outputs that are more relevant to the task.

Overall, the comparison shows that prompts can have a significant impact on the performance of LLMs for downstream applications.
Prompts can help LLMs to better understand the task at hand, focus on the most relevant information, and learn the task-specific vocabulary.
Our findings highlight the critical role of prompts in leveraging LLMs for downstream applications.
By effectively prompting LLMs, we can significantly enhance their performance across a diverse range of tasks, especially in RCA applications.


\subsection{Performance on downstream task}

\begin{table*}[ht]
    \centering
    \begin{tabularx}{15.5cm}{X | X | c c c c c} 
        \hline
        \textbf{LLM} & \textbf{Method} & \textbf{Cosine similarity} & \textbf{BLEU} & \textbf{ROUGE} & \textbf{METEOR} & \textbf{BERT} \\
        \hline
        \multirow{3}{*}{\textbf{Gemma-7B-Instruct}} & Fine-tuned & 0.544 & 0.122 & 0.183 & 0.128 & 0.843 \\ 
        
        & RAG & 0.547 & 0.181 & 0.202 & \textbf{0.303} & 0.789 \\

        & \textbf{Hybrid} & \textbf{0.708} & \textbf{0.218} & \textbf{0.316} & 0.261 &\textbf{ 0.884} \\
        
        \hline
        \multirow{3}{*}{\textbf{Llama3-8B-Instruct}} & Fine-tuned & 0.591 & 0.133 & 0.174 & 0.150 & 0.845 \\
        & RAG & 0.652 & 0.210 & 0.302 & 0.252 & 0.876 \\
        & \textbf{Hybrid} & \textbf{0.729} &\textbf{0.322}& \textbf{0.452} &\textbf{0.483} & \textbf{0.905} \\

        \hline
         \multirow{3}{*}{\textbf{Mistral-7B-Instruct-v0.2}} & Fine-tuned & 0.403 & 0.043 & 0.062 & 0.108 & 0.844 \\
        & RAG & \textbf{0.758} & 0.210 & 0.348 & 0.537 & 0.933 \\
        & \textbf{Hybrid} & 0.757 &\textbf{0.211}& \textbf{0.349} &\textbf{0.538} & \textbf{0.933} \\

        \hline
         \multirow{3}{*}{\textbf{Phi-3-mini-4k-Instruct}} & Fine-tuned & 0.394 & 0.069 & 0.097 & 0.094 & 0.657 \\
        & RAG & 0.710 & 0.264 & 0.379 & \textbf{0.431} & \textbf{0.913} \\
        & \textbf{Hybrid} & \textbf{0.724} & \textbf{0.277} & \textbf{0.404} & 0.428 & 0.903 \\

        \hline
         \multirow{3}{*}{\textbf{Falcon-7B-Instruct}} & Fine-tuned & 0.298 & 0.029 & 0.050 & 0.054 & 0.717 \\
        & RAG & 0.347 & 0.066 & 0.111 & 0.122 & 0.746 \\
        & \textbf{Hybrid} & \textbf{0.531} &\textbf{0.115}& \textbf{0.219} & \textbf{0.170} & \textbf{0.783} \\

        \hline
        
    \end{tabularx}
    \caption{Performance comparison with different LLMs. \textbf{Bold} denotes the best results on the same LLM.}
    \label{tab:eval_domain_llm}
\end{table*}

\begin{table}[ht]
    \centering
    \resizebox{\columnwidth}{!}{%
    \begin{tabular}{lcccccccc} 
        \hline
        \multirow{2}{*}{\textbf{Anomaly}} & \multirow{2}{*}{\textbf{Count}}    & \textbf{Cosine}    & \multirow{2}{*}{\textbf{BLEU}}   & \multirow{2}{*}{\textbf{ROUGE}} & \multirow{2}{*}{\textbf{METEOR}} & \multirow{2}{*}{\textbf{BERT}} \\
        & & \textbf{similarity} &  &  &   \\
        \hline
        
        Type 0 & 19 & 0.721 & 0.316 & 0.426 & 0.426 & 0.876 \\
        Type 1 & 55 & 0.731 & 0.302 & 0.415 & 0.452 & 0.900 \\
        Type 2 & 11 & 0.622 & 0.312 & 0.471 & 0.525 & 0.913 \\
        Type 3 & 44 & 0.689 & 0.301 & 0.452 & 0.481 & 0.897 \\
        Type 4 & 41 & 0.813 & 0.371 & 0.490 & 0.528 & 0.925 \\
        Type 5 & 14 & 0.779 & 0.350 & 0.431 & 0.462 & 0.905 \\
        Type 6 & 7 & 0.724 & 0.276 & 0.467 & 0.463 & 0.924 \\
        Type 7 & 3 & 0.967 & 0.690 & 0.860 & 0.904 & 0.958 \\
        Type 8 & 5 & 0.767 & 0.439 & 0.604 & 0.658 & 0.937 \\
        Type 9 & 3 & 0.764 & 0.447 & 0.663 & 0.582 & 0.939 \\
        Type 10 & 4 & 0.560 & 0.203 & 0.466 & 0.260 & 0.919 \\
        Type 11 & 8 & 0.524 & 0.177 & 0.318 & 0.355 & 0.864 \\
        Type 12 & 3 & 0.641 & 0.264 & 0.633 & 0.183 & 0.809 \\
        \hline
    \end{tabular}
    }
    \caption{Evaluating the hybrid method using the LLAMA 3 model on different anomaly types on the evaluation dataset.}
    \label{tab:eval_anomaly_type}
\end{table}


We use different models such as Gemma \cite{team2024gemma}, Llama3 \cite{llama3}, Mistral \cite{mistral2023}, Phi-3 \cite{phi3} and Falcon \cite{falcon2023} as the underlying LLMs for evaluation.
Table \ref{tab:eval_domain_llm} presents the evaluation results of our three approaches using various LLMs to build a RCA knowledge base.
In detail, we employed five distinct LLMs, namely, Gemma-7B-Instruct, LLaMA-3-B-Instruct, Mistral-7B-Instruct-v0.2, Phi-mini-4k-Instruct, and Falcon-7B-Instruct.
Indeed, LLMs trained on instructions are often designed to be adaptable to various tasks as long as clear instructions are provided \cite{PandaLM_2024}.
This characteristic is particularly relevant to our experiment and allows us to directly compare the effectiveness of different evaluation methods.
Each LLM was evaluated using three different methods: fine-tuning, retrieval-augmented generation, and a hybrid approach combining both of them.
By employing this diverse set of LLMs, we aim to provide a comprehensive evaluation of our approaches for building an RCA knowledge base, taking into account various model sizes (which fit our hardware), architectures, and resource requirements.
This approach allows us to draw more general conclusions about different method's effectiveness and identify the most suitable LLM choices for our specific application.
\begin{table*}[hbt]
    \begin{tabularx}{18cm}{X|X}
    \toprule
        \multicolumn{2}{>{\hsize=\dimexpr2\hsize+2\tabcolsep+\arrayrulewidth\relax}X}{\textbf{Anomaly}: The software module A receives packets from the network interface I by checking a completion flag F. This flag is set by the hardware H when either a small number of packets are processed or a predefined time threshold is reached. When there are intervals exceeding a specific time interval between packets, the completion flag for the delayed packet is triggered by the time-out event, resulting in a latency of approximately xxx milliseconds.} \\
    \hline
        \textbf{Root Cause:}
        The software issue in A causes product B to check if the F flag is equal to [value] and write it back to [value] when the following two conditions are met:
        \begin{enumerate}
            \item The [Protocol X] has expired.
            \item There are [number] packets that have been sent to the host memory.
        \end{enumerate}
        The [Protocol X] expiration register is set to a default value of [value] microseconds.
        If there is more than [value] interval between two packets, the [Protocol X] will expire, and the write-back for the delayed packet will be triggered, leading to around [value] latency.
        
        \textbf{Solution:}
        
        To fix this issue, the following changes can be made:
        \begin{enumerate}
            \item  Increase the [Protocol X] value to a higher value than [value] microseconds.
            \item  Send packets more frequently than every [value].
        \end{enumerate}
& 
    There is an issue with high latency in Product-vA when using Product-B :
    
    \textbf{Anomaly:}
    
    - Product-vA 2.8 + Product-B has a latency issue.
    
    - Product-vA 2.11 + Product-B has latency issue.
    
    \textbf{Root cause:}
    
    - The [Protocol X] value is set to [value] us by default.
    
    - When there is more than [value] interval between two packets, the write-back for the delayed packet will be triggered by [Protocol X] expiration, thus leading to around [value] latency.
    
    \textbf{Solution:}
    
    - The fix will be available in Product-vA 2.12 and onwards.
    
    - The [Protocol X] value can be manually modified for Product-vA 2.11.
    
    \textbf{Additional information:}
    
    - Product-vA 2.8 + Product-C has no latency issue.
    
    - Product-vA 1.10+ Product-C has no latency issue.
    
    Overall, this text describes a software issue with Product-vA and high latency when using Product-B. The issue is caused by the [Protocol X] value being set to [value] us by default and the write-back for the delayed packet being triggered by [Protocol X]  expiration. The fix for this issue will be available in Product-vA 2.12 and onwards.\\
    \end{tabularx}
    \caption{Comparison of fine-tuned method (left side) and hybrid method (right side)\protect\footnotemark[4].}
    \label{tab:example_compariasion_llm_rag}
\end{table*}

Across all LLMs, the hybrid method consistently achieved the highest performance according to the BERTScore metric, demonstrating its effectiveness in generating text that is more semantically similar and fluent compared to either fine-tuning or RAG methods alone.
Our hybrid approach combines the domain-specific weight adaptation of the fine-tuned model (D-LLM) with the dynamic context retrieval capabilities of RAG.
This enables the model to speak the \textit{language} of the network through fine-tuning while referencing the most up-to-date evidence via RAG, in contrast to a fine-tuning-only approach, which lacks access to external knowledge.
For example, Mistral-7B-Instruct-v0.2 achieved the best overall performance with the hybrid approach, attaining a BERTScore of 0.933.
This highlights the potential of the hybrid method, particularly for LLMs such as Mistral, which can benefit from both knowledge grounding and stylistic refinement.
Furthermore, the results reveal interesting patterns across different LLMs.
For instance, LLaMA-3-B-Instruct demonstrated the highest cosine similarity scores, indicating its ability to generate text that closely aligns with the reference text in terms of word usage and sentence structure.
On the other hand, Phi-mini-4k-Instruct showed strong performance in ROUGE and METEOR metrics with the RAG approach, suggesting its effectiveness in capturing factual content and key phrases from the reference text.
Upon investigating the lower performance of the fine-tuned-only approach, we observed that it lacks access to external domain knowledge beyond the training data, limiting its ability to generalize to rare network anomalies and resulting in hallucinated solutions.
In contrast, RAG and hybrid methods can dynamically retrieve relevant context, enabling them to handle diverse and specialized terminology more effectively.
Consequently, the outputs of the fine-tuned model are often more constrained and less semantically accurate or complete for certain RCA tasks.

To gain deeper insights into the performance of the proposed hybrid method for RCA applications, we conducted experiments across all network anomaly types.
As presented in Table \ref{tab:eval_anomaly_type}, 
the results highlight the hybrid method's performance across 13 distinct anomaly types within our testing system.
For anomaly types with larger ticket counts (e.g., types 1, 3, and 4), the metrics demonstrate a balanced performance, e.g., METEOR scores ranging from 0.452 to 0.528, BERTS scores consistently around 0.9 and cosine similarity around 0.7.
This suggests that the hybrid method is well-suited to handling diverse datasets effectively.
In contrast, anomaly types with fewer tickets, such as types 0, 2, 5, 6, 7, 8, and 9, exhibited strong performance, with metric values reaching approximately 0.4–0.6 for METEOR, 0.7 for cosine similarity, and 0.9 for BERTScore.
These results further highlight the hybrid method’s ability to generalize effectively, even when trained on limited data.
However, lower scores were observed for anomaly types such as 10, 11, and 12 (e.g., METEOR = 0.183 for type 12), indicating challenges in capturing the semantic nuances specific to these categories. It is important to note that these anomaly types had relatively small ticket counts, which likely influenced the observed performance.
Additionally, upon investigating the contents of these anomaly types, we found that the terminology associated with them is highly specialized and less frequent in the fine-tuning data.
This significant variance in terminology, along with the lack of exposure to such terms during fine-tuning, contributes to the difficulty in capturing the underlying patterns accurately.
For instance, specific terminal commands related to specific settings or error investigations, or "error\_flags" keywords defined by software, are not only unique but also rarely shared across other anomaly types, making them particularly challenging for the model to generalize effectively.
Although the model generalizes well for common anomalies, future work should target rare types using data augmentation, expert-in-the-loop mechanisms (as an option in Figure 3), or synthetic cases to reduce hallucinations and improve root-cause attribution.

Please note that, in our evaluation, we prioritize the METEOR metric over ROUGE and BLEU, as the former considers synonyms (e.g., "Node" and "Server"), while they are based on n-gram overlap methods.
This distinction is particularly significant in RCA tasks where the use of synonymous terms is common.

As the result of constructing rules for RCA and SA, the hybrid approach demonstrably produced clear and well-defined rules.
Notably, these rules exhibited a high degree of concordance with the system's product and the specific software versions currently operational within our system, refer to Table \ref{tab:example_compariasion_llm_rag} for a detailed comparison.
Furthermore, the configuration of a similarity threshold within the hybrid approach allows for the clustering of similar network anomalies.
As stated in Table \ref{tab:result_group_tickets}, the number of rules can be compressed significantly by using similarity thresholds.
This process facilitates the identification of recurring root causes and equivalent solutions.
By grouping anomalies based on their similarity, we can gain deeper insights into the underlying issues and develop more comprehensive mitigation strategies, an example of compressing five similar anomalies is illustrated in Table \ref{tab:example_on_compress_rules}.

\begin{table}[hbt]
    \centering
    \begin{tabularx}{7cm}{c | c c} 
        \hline
        \textbf{Similarity threshold} & \textbf{Total rules} & \textbf{Rule reduction} \\
        \hline

        90\% & 1048 & 1\% \\
        80\% & 1035 & 1.33\% \\
        70\% & 930 & 11.34\% \\
        60\% & 620 & 50.43\% \\
        
        \hline
    \end{tabularx}
    \caption{Number of rules compressed by similarity thresholds.}
    \label{tab:result_group_tickets}
\end{table}
\begin{table*} [hbt]
    \begin{tabularx}{18cm}{X|X|X}
    \toprule
        \multicolumn{3}{>{\hsize=\dimexpr3\hsize+3\tabcolsep+\arrayrulewidth\relax}X}{
        \textbf{Anomaly}: network disruptions due to CARD-S1 and CARD-S2 failures.
        
        \textbf{Insights}

        - CARD-S1 failures can cause CARD-S2 to go offline, leading to endpoint outages and traffic disruptions.

        - CARD-S2 switchovers can happen abruptly due to failed CARD-S2 cards, causing temporary service interruptions.
 
        \textbf{Logs}
        
        - Persistent Log: "CARD-S1 set severity and trigger reason: Fault Severity change on Secondary. No Faults $-->$ Card Degrade due to fault XXX-ALARM1 (10402)"
        
        - Persistent.log: "YYMMDD HHMM: \%PAD-6-INFO: slot CARD-S1-1, ALARM\_MAJOR: Hardware Failure YYMMDD HHMM: \%PAD-6-INFO: slot CARD-S1-2, ALARM\_MAJOR: Hardware Failure"
        
        - etc
        
        }\\
        \hline
        Root cause & Products & Solution \\
        \hline \\
        Mastership synchronization was lost for CARD-S2 & Product-A, Product-B 8000 4 & This is a software issue and fix will be delivered from Product-C 4.2 DP 11 release. Fix will be delivered from Product-C 4.2 DP 11 release \\ [20pt]
        
        Hardware failure & Product-A 1, Product-C, Product-E, Product-E 1.10 & Replace the failed card, check the card for any physical damage or wear and tear. \\ [20pt]

        CARD-S1 switch-overs happened frequently due to GESW link issue between CARD-S2-1$->$CARD-S11 and CARD-S2-1$->$CARD-S1-2 & Product-A 1 & This is a Hardware Issue. Link refer: [System-inter-linl-was-replaced] \\
        \hline
    \end{tabularx}
    \caption{An example of compress 5 similar anomalies based on a threshold of 70\% \protect\footnotemark[4].}
    \label{tab:example_on_compress_rules}
\end{table*}
Thus, this approach not only improves efficiency by reducing the workload on the D-LLM for our application but also provides D-LLM with a holistic understanding of the network anomaly and root causes.
However, it is important to note that there is a trade-off between efficiency and accuracy.
By using a small similar threshold to compress rules, there is a risk of missing some important information.
The optimal similarity threshold will depend on the specific application and its requirements.
Another notable concern is the increased computational cost compared to the fine-tuning approach.
The retrieval process in RAG and hybrid methods adds an extra layer of complexity, requiring efficient vector database management and similarity search algorithms.
During our experiments, we observed that the response time for similar inputs in both the RAG and hybrid approaches increased by approximately four seconds, with the majority of the delay attributable to the retrieval process when the procedure queried relevant information.
In contrast, the average GPU memory consumption remained largely consistent across approaches.
However, these performance observations may vary depending on the type of GPU, the size of the database and the retrieval algorithms employed.

Recent literature emphasizes knowledge graph (KG) augmentation for RCA and service assurance, for example, in Wang \textit{et al}. \cite{wang2024kgroot}.
These approaches are shown to be effective when RCA data is structured with well-defined relationships between entities such as network elements, alarms, root causes and solutions.
However, in many real-world telecom operations, the primary artifacts for RCA are support ticket corpora, which mainly consist of unstructured text, expert narratives, and solution log formats that have been in practical use for years and do not readily map onto graph structures.
Consequently, our method focuses on a hybrid approach that directly leverages these unstructured ticket datasets without additional graph engineering.
Our results show that {\telcomodel} can effectively capture context and perform generalizable RCA by linking relevant historic evidence in ticket corpora, yielding accuracy and semantic consistency competitive with KG-based methods, particularly when the underlying data is primarily unstructured.
Furthermore, to validate the correctness of the results, a subset of the generated RCA rules was reviewed by our industrial partner, and their correctness and actionability were confirmed.

Ultimately, our results demonstrate that the hybrid approach, combining fine-tuning and RAG, offers a promising direction for enhancing the performance of LLMs in automatically building a knowledge base for RCA.
Its superior performance across multiple metrics and different LLMs makes it the most effective method for constructing knowledge bases in our target application.

\section{Conclusion}
\label{sec:conclusion}

In this study, we propose {\telcomodel}, a method for automating the construction of a knowledge base for RCA, and evaluate three approaches for building such a knowledge base using support tickets tailored to RCA and service assurance (SA).
These approaches were evaluated using various metrics for both vocabulary and semantic similarity.
The fine-tuned LLM showed promising results in lexical similarity (Cosine similarity, BLEU, METEOR, and ROUGE) but lacked fluency and semantic understanding (BERT).
RAG augmentation improved these aspects but the hybrid approach excelled across almost all metrics evaluated on multiple LLM models.
The hybrid method aligns with real-world scenarios, generating rules that match specific systems, products and software versions.
Furthermore, similarity thresholds within the hybrid approach enabled efficient clustering of similar network anomalies, improving the identification of root causes and mitigation strategies.
However, the trade-off between efficiency and accuracy inherent in rule compression requires further exploration.
We acknowledge that our evaluation is based on a dataset that, while reflecting real-world operational challenges, is specific to the telecommunications domain and relatively limited in scale. 
Although effective for specialized domain adaptation, extending the approach to larger, multi-domain environments may require additional data augmentation strategies to mitigate the scarcity of rare anomaly classes.
Overall, our study highlights the effectiveness of the hybrid approach for building knowledge bases in RCA and service assurance tasks.

In future work, we plan to address the limitations of this study and explore several key areas.
First, to enhance the scalability and generalizability of {\telcomodel}, we will evaluate its performance on larger, more diverse datasets from different industrial domains.
To mitigate performance drops for rare anomaly types, we plan to explore advanced solutions for data imbalance.
This includes generating synthetic data and leveraging paraphrasing and back-translation to expand low-frequency classes. 
Additionally, we plan to explore different prompt engineering techniques, where rare-category exemplars guide model inference, together with adaptive loss weighting during fine-tuning to further enhance performance.

\bibliographystyle{IEEEtran}
\bibliography{2nd_journal_refs,BJ}


\begin{IEEEbiography}[{\includegraphics[width=1in,height=1.25in,clip,keepaspectratio]{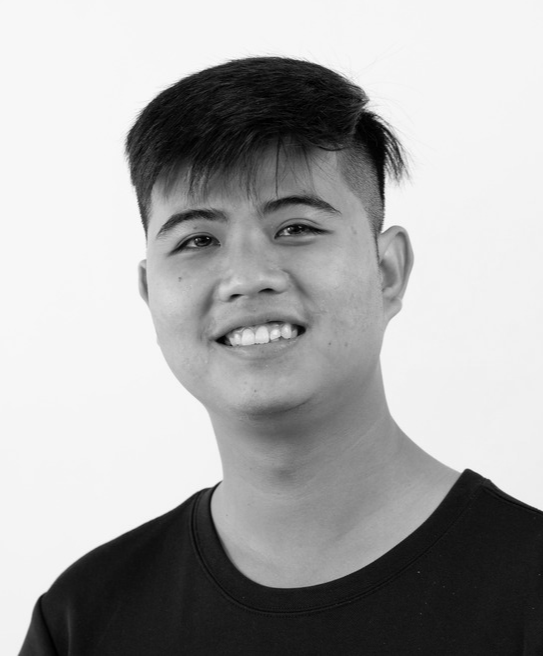}}]{Nguyen Phuc Tran}~received his M.S. degree in Computer Science from the University of Information Technology, Vietnam National University, Ho Chi Minh City, in 2020. Since 2021, he has been pursuing his Ph.D. at Concordia University, Montreal, Quebec, Canada. With over five years of experience as a senior software engineer in system development and telecommunication technology, he has honed his expertise in system optimization, security, quality assurance, data analysis, team leadership, and stakeholder management. His current research interests encompass the design and application of artificial intelligence, including large language models, in mobile communication networks. He focuses on areas such as resource allocation, energy efficiency, green mobile networks, system design, root cause analysis, and system optimization.
\end{IEEEbiography}

\begin{IEEEbiography}[{\includegraphics[width=1in,height=1.25in,clip,keepaspectratio]{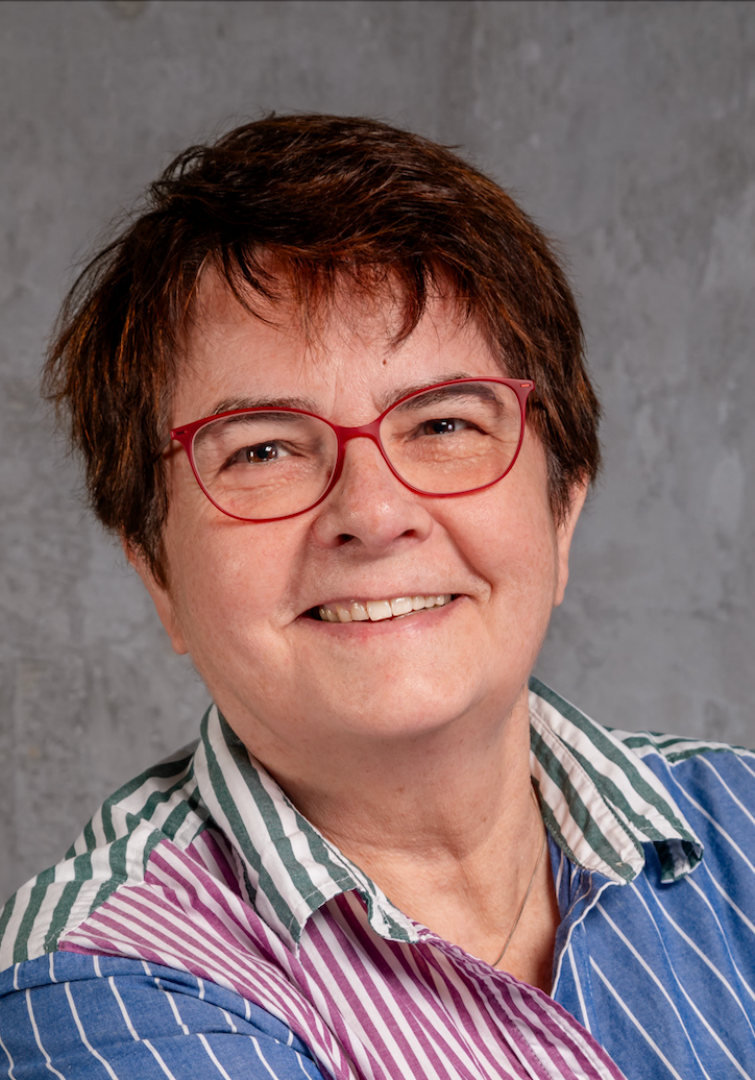}}]{Brigitte Jaumard}~(Senior Member, IEEE)~is the scientific director of Confiance IA, an Industrial research consortium - trustworthy AI supported by the Quebec government. She is also a professor in the Computer Science and Software Engineering (CSE) Department at Concordia University. Her research focuses on mathematical modelling and algorithm design (large-scale optimization and machine learning) for problems arising in communication networks, transportation and logistics networks. 
Recent studies include the design of efficient optimization/machine learning algorithms for network design, dimensioning and provisioning, scheduling in edge-computing and clouds, and 5G networks. During her 2020-2021 sabbatical year, she was a senior advisor for the Montreal Ericsson GAIA (Global Artificial Intelligence Accelerator) research center and the chief scientist of CRIM.

Brigitte Jaumard was ranked among the top 2\% of scientists in her field of research according to a 2021 study based on research citations. She was awarded several research chairs (Canada Research Chair and Concordia Research Chair, both Tier I during the years 2000-2019). B. Jaumard has published over 300 papers in international journals in Operations Research and in Telecommunications.

\end{IEEEbiography}

\begin{IEEEbiography}[{\includegraphics[width=1in,height=1.25in,clip,keepaspectratio]{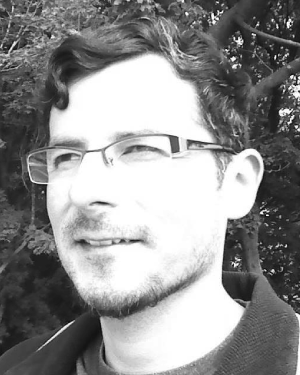}}]{Oscar Delgado}~received the Ph.D. degree in electrical engineering from McGill University, Montreal, in 2016. 
After his Ph.D., he was a postdoc researcher at the Telecommunications and Signal Processing Laboratory (TSP), department of Electrical and Computer Engineering, McGill University.
He is currently a research associate at \'{E}cole the Technologie Sup\'{e}rieure (ETS)
His research interests include but are not limited to applications of 5G wireless mobile communication technologies, AI/ML, network virtualization, service assurance, green wireless systems, analysis and design of video traffic management techniques, resource allocation strategies, and energy efficiency algorithms.
\end{IEEEbiography}

\begin{IEEEbiography}[{\includegraphics[width=1in,height=1.25in,clip,keepaspectratio]{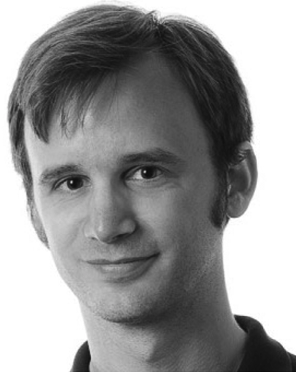}}]{Tristan Glatard}~is an associate professor with the Department of Computer Science and Software Engineering, Concordia University in Montreal, and Canada research chair (Tier II) on Big Data Infrastructures for Neuroinformatics. Before that, he was research scientist with the French National Centre for Scientific Research and visiting scholar with McGill University.
\end{IEEEbiography}

\begin{IEEEbiography}[{\includegraphics[width=1in,height=1.25in,clip,keepaspectratio]{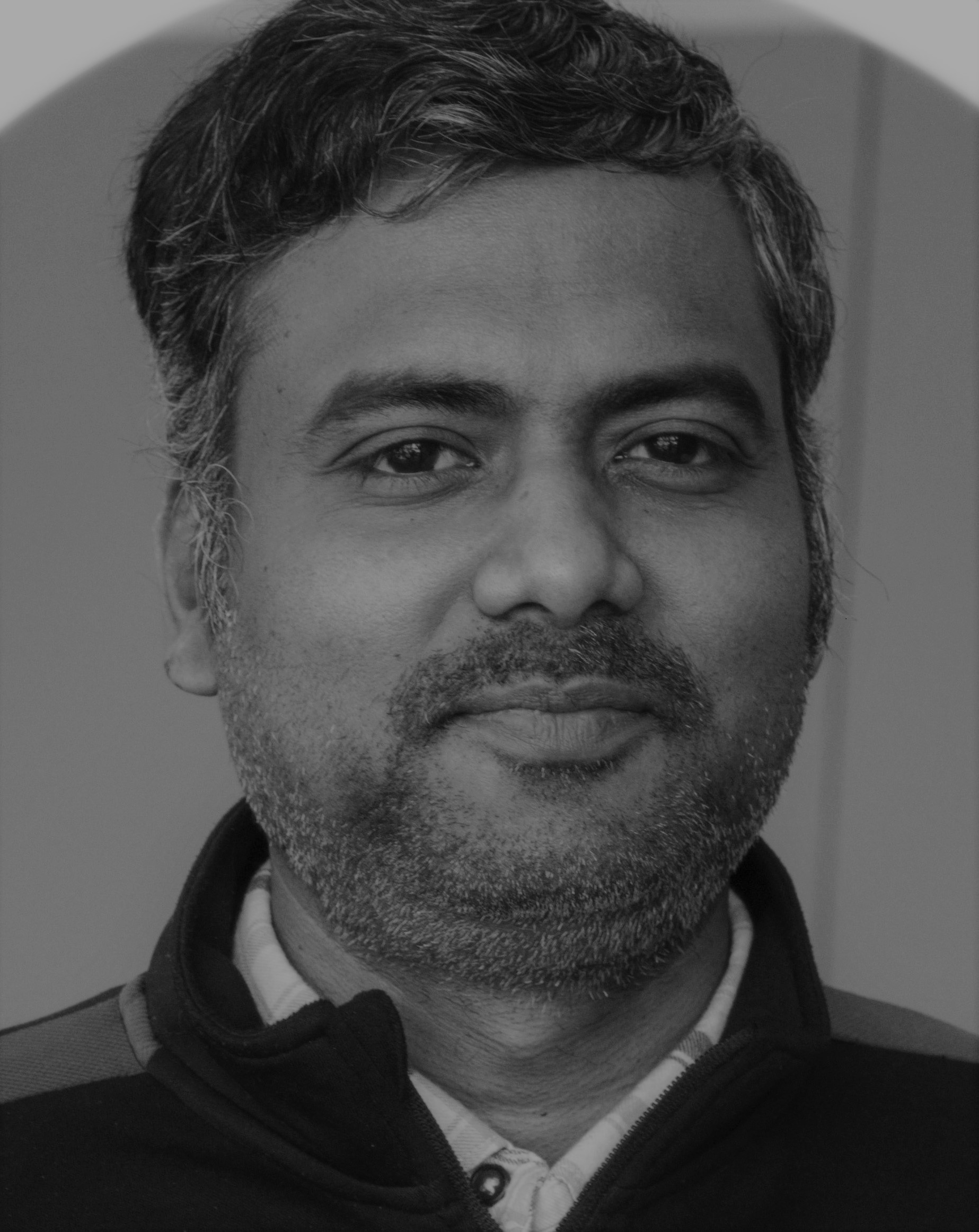}}]{Karthikeyan Premkumar}~holds a Bachelor's degree in Engineering from Madras University, India, and is currently pursuing a Master of Science degree at McGill University in Canada. With over 18 years of experience in the Telecom industry, he specializes in industrializing AI/ML solutions and designing System \& Solution architectures in BSS, Cloud platforms, and Infrastructure platforms for Telecom operator networks. Presently, he serves as a Principal Data Scientist, leading the design of Knowledge Models and Machine Reasoning technologies for cognitive networks. He has a notable publication record, including over 10 patents and papers on Telecom analytics.
\end{IEEEbiography}

\begin{IEEEbiography}[{\includegraphics[width=1in,height=1.25in,clip,keepaspectratio]{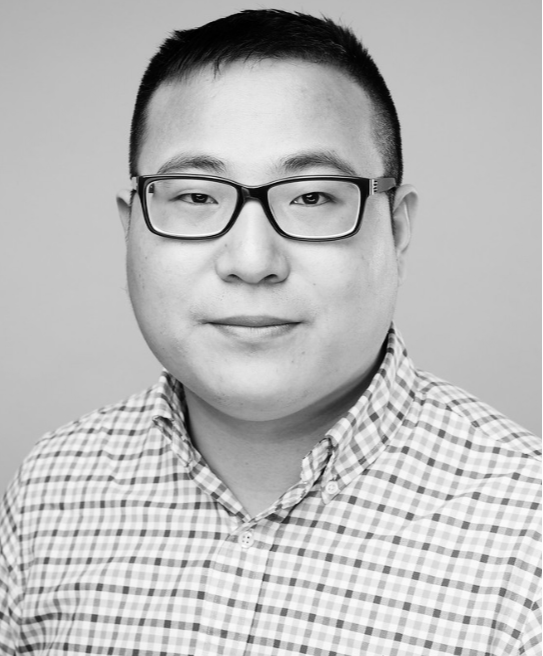}}]{Kun Ni}~received the M.S. degree in Computer Science from University of Montreal, Canada and an undergraduate degree in Software engineering from Southeast University, China.
Currently, he is a Data scientist and Machine Learning Developer at Ericsson.
As a passionate AI/ML developer, he has worked on various projects, including predictive modelling, time series analysis, and natural language processing. Proficient in Python, C++, TensorFlow, and PyTorch.
\end{IEEEbiography}

\end{document}